\documentclass[letterpaper, 10 pt, conference]{ieeeconf}  
\usepackage{microtype}
\usepackage{graphicx}
\usepackage{subfigure}
\usepackage{booktabs}
\usepackage{subfigure}
\usepackage{epsfig}
\usepackage[labelfont=bf]{caption}
\usepackage{indentfirst}
\usepackage{amsmath}
\usepackage{amssymb}
\usepackage{mathtools}
\usepackage{footmisc}
\usepackage{xcolor}
\usepackage[pagebackref,breaklinks,colorlinks]{hyperref}
\usepackage{hyperref}
\usepackage[capitalize,noabbrev]{cleveref}
\usepackage{float}
\IEEEoverridecommandlockouts
\title{\LARGE \bf
Multi Agent Navigation in Unconstrained Environments using a Centralized Attention based Graphical Neural Network Controller
}

\author{Yining Ma*$^{1,2}$, Qadeer Khan*$^{1,2}$ and Daniel Cremers$^{1,2,3}$% <-this % stops a space
\thanks{$^{*}$ These authors contributed equally.}
\thanks{$^{1}$ All authors are affiliated with the Computer Vision Group, School of Computation, Information and Technology, Technical University of Munich. Contact:        {\tt\small \{yining.ma, qadeer.khan, cremers\}@tum.de}}%
\thanks{$^{2}$ All authors are affiliated with the Munich Center for Machine Learning (MCML).}
\thanks{$^{3}$ The author is affiliated with the University of Oxford.}
}

\begin{document}

\maketitle
\thispagestyle{empty}
\pagestyle{empty}

%%%%%%%%%%%%%%%%%%%%%%%%%%%%%%%%%%%%%%%%%%%%%%%%%%%%%%%%%%%%%%%%%%%%%%%%%%%%%%%%
\begin{abstract}

In this work, we propose a learning based neural model that  provides both the longitudinal and lateral control commands to simultaneously navigate multiple vehicles. The goal is to ensure that each vehicle reaches a desired target state without colliding with any other vehicle or obstacle in an unconstrained environment. The model utilizes an attention based Graphical Neural Network paradigm that takes into consideration the state of all the surrounding vehicles to make an informed  decision. This allows each vehicle to smoothly reach its destination while also evading collision with the other agents. The data and corresponding labels for training such a network is obtained using an optimization based procedure.  Experimental results demonstrate that our model is powerful enough to generalize even to situations with more vehicles than in the training data. Our  method also outperforms comparable graphical neural network architectures. Project page which includes the code and supplementary information can be found here: \href{https://yininghase.github.io/multi-agent-control/}{https://yininghase.github.io/multi-agent-control/}

\end{abstract}

\section{INTRODUCTION} \label{sec:introduction}

\begin{figure}[!h]
\centering
\includegraphics[width=\columnwidth]{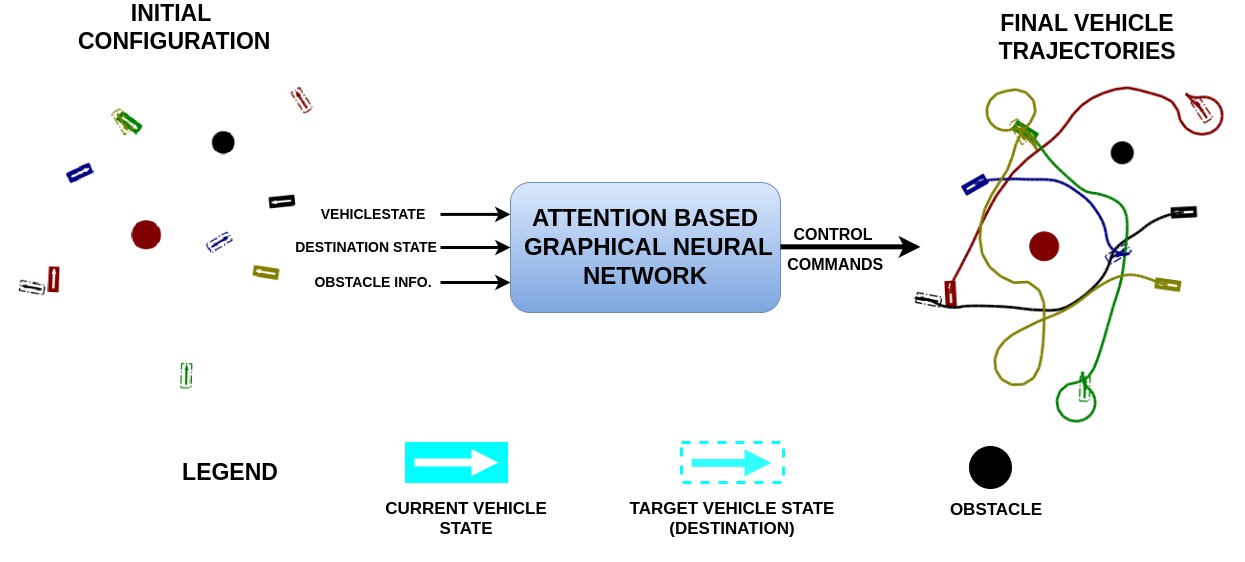}
\caption{\label{fig:overview} \textbf{Overview of multi-agent control:} The initial configuration (on the left) shows five vehicles colored \textbf{black}, \textbf{\textcolor{red}{red}}, \textbf{\textcolor{green}{green}}, \textbf{\textcolor{blue}{blue}} and \textbf{\textcolor{olive}{olive}}. The initial starting state of the vehicles is represented by a rectangle with solid boundaries. The arrow within each rectangle depicts the corresponding orientation of that vehicle. Meanwhile, the rectangles with broken boundaries represents the desired destination/target position of each vehicle. We would like to produce the sequence of control actions such that the five vehicles safely reach their destination state without colliding with each other or the circled obstacle. These control actions are produced by the Attention Based Graphical Neural Network (A-GNN). The A-GNN receives information about the current state and desired destination state of all the five vehicles along with information about any obstacle. The network outputs the control commands for all the five vehicles together. These control commands are executed for all the five vehicles simultaneously. Each vehicle then attains a new state. This new state is then fed again to the A-GNN as the current state to predict the new steering command. This process is iteratively repeated until all the vehicles reach their corresponding destination state.  The trajectory traversed by all the vehicles as a result of this iterative process is shown on the right of the figure. Some video examples of our model demonstrating this can be found: \href{https://yininghase.github.io/multi-agent-control\#Results-of-our-Model}{here} }
\end{figure}

 Data driven approaches to senorimotor control have seen a meteoric growth with the advent of deep learning  in the last decade \cite{rausch2017learning,eraqi2017end,levine2016end,sobh2018end}. Powerful neural network architectures can now be trained and deployed in real-time applications \cite{nvidiaend2end}.  The recent success of deep learning for agent control can primarily be attributed to the following 2 factors:
\begin{enumerate}
    \item Cheap hardware accelerators that exploit the parallel computations\cite{perez2019low}, particularly in deep neural architectures \cite{8977867}.
    \item The availability of simulation platforms that allow benchmarking and evaluation of various vehicle control algorithms \cite{shah2018airsim,dosovitskiy2017carla}.
\end{enumerate}

\cite{9812316,9863660} have used such platforms to evaluate their learning based control algorithms. However, many learning based control approaches have certain limitations of their own:
\begin{enumerate}
    \item They require collection of tremendous amounts of labeled supervised data for training which in some cases may not even be available. For e.g. recovering a vehicle from driving on a sidewalk.
    \item The neural network is trained to  control only one vehicle. Moreover, since the sensors are placed on the ego-vehicle, the model has partial observability of the environment. Therefore, traffic rules are used to regulate the flow of vehicles and prevent any untoward incident. Moreover, for the task of autonomous driving, the vehicle is constrained to drive only on the road.
\end{enumerate}

In this work, we present a technique to control not one but a variable number of vehicles in an unconstrained environment having obstacles. The vehicles are meant to reach their desired destination/target state without collision. Supervised labeled data is also not available. Rather, we optimize against a cost function to determine the longitudinal and lateral control labels. This optimization procedure for the task of label generation is only done offline and hence does not influence the real-time operation.

Figure \ref{fig:overview} describes this process. The Attention based Graphical Neural Network (A-GNN) is a core component which makes these control command predictions. It takes information of all obstacles,  the current state of all vehicles and their desired destination/target state. The output of the A-GNN are the control commands for all the vehicles. These control commands are then executed which results in the vehicle achieving a new state. This new state along with the target/desired destination state of the all the vehicles is fed back to the A-GNN to produce the new control commands. The process is iteratively repeated until all vehicles reach their destination state.  Note that the A-GNN is trained with labels obtained from offline optimization performed against a cost function.

Note that the A-GNN allows each vehicle to attend to the information of other vehicles. This allows each vehicle to make informed control decisions in order to avoid collisions among themselves while also successfully reaching their destination. 

We summarize the contributions of our framework below:
\begin{enumerate}
    \item Ability to control a variable number of vehicles to reach their desired destination states. We show that our model can even perform inference for more number of vehicles than for which it was initially trained.
    \item The architecture of our A-GNN outperforms other comparable attention based GNN layers such as GAIN \cite{9827155}, TransformerConv \cite{shi2020masked}, non-attention based architectures such as EdgeConv \cite{yue2019dynamic} and also the naive multi-layer percepton (MLP) method that does not utilize any edges in the graphical structure.
    \item Our A-GNN can handle the presence of both dynamic vehicles and static obstacles.
    \item We have released the implementation of our framework here: \href{https://github.com/yininghase/multi-agent-control}{https://github.com/yininghase/multi-agent-control}.
\end{enumerate}

An application of this work, could be in large unmanned indoor warehouses where multiple agents are transporting items from one location to another. We assume that all the agents/vehicles move in the x-y plane and can be localized to determine their state. Localization in an indoor environment can be done using for eg. the approaches from \cite{8767421,wang2016csi}.

\section{Related Work}\label{sec:related_work}

\noindent{\textbf{Multi-Agent Trajectory Prediction \& Control:}} \cite{yuan2021agentformer,tang2022evostgat,ivanovic2019trajectron} model pedestrian behaviour to predict their future trajectory primarily utilizing information from the past. Our model is rather concerned with control of vehicle agents using current state and destination information. \cite{MOORTHY2022308,HE2022651} do focus on multiple agents but for the task of leader guided formation control. Meanwhile, \cite{DUTTA202333} investigates connectivity restoration for formation control. In our work, the multiple vehicles are neither guided to follow a leader nor are they in pursuit of a formation. Rather, the objective of each vehicle is to independently reach its desired destination while taking information of other vehicles into consideration. Co-operation among the vehicles in our framework only exists to the extent that each vehicle can reach its desired destination without collision with each other and the obstacles. \\

\noindent{\textbf{Attention based architectures:}} In the context of deep learning, the attention mechanism was primarily introduced in the discourse of Natural Language Processing \cite{bahdanau2014neural}. However, it has gained utility in other domains too such as for the task of single vehicle control \cite{Chitta2021ICCV}. Here, attention is applied between different patches of the sensory data for the same vehicle. In contrast, our method applies attention  between the different vehicles to be controlled. \cite{9827155} uses an  Graph Attention Isomorphism Neural Network (GAIN) architecture, where the task is to predict the trajectory of multiple vehicles using their past trajectory as input. Our method in contrast predicts the control commands using only the existing and destination state rather than the entire historical trajectory. \\

\noindent{\textbf{Reinforcement Learning}}
\cite{8917268,9151648} use reinforcement learning for navigating vehicular-traffic. However, this is done by controlling the traffic signal rather than controlling the individual vehicles. Moreover, this is done only at intersections where clear rules for vehicle navigation exist. In our case, we handle an unconstrained environment where no predefined rules exist for negotiating bottlenecks. The only objective that the vehicles must ensure is that they avoid collisions while reaching the destination. \cite{haworth2020deep} also use RL in simulation but for the control of humanoids rather than vehicles. \cite{sartoretti2019primal} performs multi-agent path finding but in a discrete action space in the gridworld. This is as opposed to our approach which is in continuous action space and environment. The co-operative navigation task of \cite{lowe2017multi} is similar to our task of navigating multiple agents to desired destinations. However, the agents are treated as particles while our approach considers the vehicle kinematics. Another issue with RL methods is that the training tends to be heavily sample-inefficient \cite{electronics9091363,8686348}  requiring far greater training sessions than methods where the target labels are already known such as in our case through optimization techniques. \\

\noindent{\textbf{BEV representation:}}
Note that the state of the vehicles are represented in a Bird's Eye View (BEV) format. This format has extensively been used in various perception related tasks such as object detection, tracking, segmentation  \cite{Peng_2023_WACV,luo2018fast} etc. It has also demonstrated to be very convenient for planning and control tasks. For e.g. \cite{bansal2018chauffeurnet} takes the BEV representation along with the intention of a single vehicle as input to predict its trajectory. \cite{chen2019lbc} also uses the BEV representation to first train a teacher for the task of single vehicle control. Knowledge distillation is then used to train a subsequent student model that takes image data as input for control prediction.  \cite{Zhang_2022_CVPR} uses pseudo-labeling of web images as part of a pipeline that predicts future waypoints in the BEV space. The aforementioned approaches for prediction and control are only tailored towards single vehicles. Our approach on the other hand is capable of handling multiple vehicles. 

\section{Framework}\label{sec:framework}

In this section, we first describe the architecture of our Attention based Graphical Neural Network (A-GNN) (See Subsection \ref{subsec:A-GNN}). Next, the process to generate the steering labels for training the A-GNN is discussed (See Subsection \ref{subsec:label_gen}).

\subsection{Attention Graph Neuron Network}\label{subsec:A-GNN}

The A-GNN takes as input the state information of both the $N_{vehicle}$  dynamic vehicles and $N_{obstacles}$ static obstacles in the scene to predict the control commands (steering angle, $\varphi$ and pedal acceleration, $p$) for all vehicles. Each vehicle/obstacle is indexed by $i$ $\in$ [ 1, $N_{vehicle}+N_{obstacle}$]. The feature vector for entity $i$ at layer $l$ of the neural network is denoted by $z^{l}_{i}$. The input feature vector for each entity is given by $z^{0}_{i} \in \mathbb{R}^{8}$ representing the current location ($x$ and $y$), current orientation ($\theta$), current velocity $v$, target position ($\hat{x}$ and $\hat{y}$), target orientation ($\hat{\theta}$) and whether or not the entity is a vehicle (0) or a circular obstacle (with radius $r$). Hence, with this representation 
$z^{0}_{i_{vehicle}} = [x, y, \theta, v, \hat{x}, \hat{y}, \hat{\theta}, 0]^{T}$. Meanwhile, $z^{0}_{i_{obstacle}} = [x, y, \theta, 0, x, y, \theta, r]^{T}$. Note that the target state of the static obstacle is the same as its current state. This is because the obstacles are stationary.

We now construct a Graph wherein each vehicle and obstacle is considered a node. Input node $i$ is represented by the feature vector $z^{0}_{i}$. Note that each vehicle node in the graph needs to retrieve state information about all other entities in the environment in order to avoid collision and reach its desired destination. Therefore, we build an edge from all the other vehicle nodes and all obstacle nodes towards this vehicle node. Meanwhile, the obstacle nodes are static and do not have any incoming edge. Mathematically, for any vehicle node $i$ its neighbors in the Graph $G$  are $N_{i} = \{ j | j = 0,1,2...,N_{vehicles}+N_{obstacle} \cap j \neq i \}$ and for any of the obstacle nodes its neighbor in the graph $G$ is $\emptyset$. 

This Graph $G$ is then passed through a series of neural layers to eventually predict the control command for each vehicle. The control command $\in \mathbb{R}^{2}$ corresponds to the steering angle ($\varphi$) and pedal acceleration ($p$) of the vehicle. The flow of information through the A-GNN is described as follows:

The input node features are first converted to a higher dimensional latent vector: 
$z^{1}_{i} = \sigma^{1}(W^{1} \cdot z^{0}_{i}) \in \mathbb{R}^{d_{1}}$, where $W^{1}$ $ \in \mathbb{R}^{d_1 \times 8}$ is a trainable weight matrix, while  $\sigma^{1}$ is the non linear ReLU activation for the first layer.  
Next, a series of $L$ residual graphical layers are used. In these layers information about the neighbouring entities is retrieved by each node via an attention based mechanism. The residual connection carries information from the preceding layer $l-1$ to a successive layer $l+1$. This ensures that prior information important for the network is carried forward.  Mathematically, this process is described by:
\newline
\newline
\noindent for $k = 1,2,...,{L}$: 
\begin{equation}
\begin{split}
z^{2k}_{i} & = \sigma^{2k}(F^{2k}_{Att}(z^{2k-1}_{i}, z^{2k-1}_{N_{i}})) \\
z^{2k+1}_{i} & = \sigma^{2k+1}(F^{2k+1}_{Att}(z^{2k}_{i},   z^{2k}_{N_{i}}) + z^{2k-1}_{i}) 
\end{split}
\end{equation}
where $z^{2k-1}_{i} \in \mathbb{R}^{d_{2k-1}}$, $z^{2k}_{i} \in \mathbb{R}^{d_{2k}}$ and $z^{2k+1}_{i} \in \mathbb{R}^{d_{2k+1}}$. To cater for the residual connection in the last equation of the for loop, note that: $d_{2k+1} = d_{2k-1}$.
Meanwhile $F^{l+1}_{Att}$ describes the \emph{Attention} mechanism at layer $l+1$ and is defined by: ,  
\begin{equation}\label{eq:Fatt}
  F^{l+1}_{Att}(z^{l}_{i},   z^{l}_{j})) = W^{l}_{self} \cdot z^{l}_{i} + \sum_{j \in N_{i}} \alpha_{ij} \cdot F_{value}(z^{l}_{i} | z^{l}_{i}-z^{l}_{j}))  
\end{equation}
\newline
where $\alpha_{ij} = {{F_{query}(z^{l}_{i})^{T} \cdot F_{key}(z^{l}_{j})} \over { \sum_{k \in N_{i}} F_{query}(z^{l}_{i})^{T} \cdot F_{key}(z^{l}_{k})}}$ are the attention weights of the neighbours of vehicle $i$.  
Meanwhile, $F_{value}$, $F_{query}$, $F_{key}$ are the joint trainable parameters of a U-Net inspired architecture with shared encoder weights for the attention mechanism. Details of this U-Net inspired attention mechanism can be found in Subsection \ref{subsec: U-Attention origin}.

The final layer $F$ of the A-GNN outputs the steering commands for all the vehicles:
\begin{equation}
z^{F}_{i} = \sigma^{F}(W^{F} \cdot z^{2L+1}_{i})
\end{equation}
$\sigma^{l}$ with $l \in \{1,...,2L+1\}$ is the ReLU non-linear activation. Meanwhile, $\sigma^{F}$ is chosen to be the scaled \emph{tanh} function to accommodate negative values of the steering angles and reverse pedal acceleration. 

\subsection{Label Generation Process} \label{subsec:label_gen}
In this subsection, we describe the procedure for generating the data and corresponding control commands. This is done by first creating a motion model of each vehicle using the kinematic bicycle model \cite{933130}. Next, the optimization is run such that all vehicles reach their desired destination state without colliding with each other or any obstacles in the scene. Once we generate enough data, the A-GNN model architecture described in Subsection \ref{subsec:A-GNN} is trained to predict the control commands for the corresponding vehicle scenarios in the dataset. Now one may rightfully ask, what is the advantage of training the A-GNN when we can already generate the control commands from optimization? The reason is that as number of cars in the environment is increased, the optimization becomes extremely slow and may not be applicable for real time operation. The advantage of the A-GNN is that it learns to extract patterns from the data and applies them to similar situations not seen during training. In fact, in the experiments section, we show that the data is generated using optimization for a maximum of 3 vehicles in the environment. However, the A-GNN is powerful enough to even make predictions for controlling 6 vehicles in the scene. Meanwhile, please refer to Subsection \ref{subsec:runtime} for a comparison of run times, wherein the inference time of our model remains fairly stable with the increase in the number of vehicles/obstacles. Recall that the state of the vehicle is described by its location ($x$ and $y$), orientation/angle ($\theta$) and velocity ($v$). The vehicle can be controlled by adjusting the acceleration from the pedal ($p$) and maneuvering the steering angle ($\varphi$). 
Using the kinematic bicycle model, the equations of motion can be updated from time $t$ to $t+1$ using the following:
\begin{equation}
\begin{split}
x_{t+1} & = x_{t} + v_{t} \cdot \cos(\theta_{t}) \cdot \Delta t \\ y_{t+1} & = y_{t} + v_{t} \cdot \sin(\theta_{t}) \cdot \Delta t \\ \theta_{t+1} & = \theta_{t} + v_{t} \cdot \tan(\varphi_{t}) \cdot \gamma \cdot \Delta t \\
v_{t+1} & = \beta \cdot v_{t} +  p \cdot \Delta t 
\end{split}
\end{equation}
where, $\beta$ and $\gamma$ are tuneable parameters during optimization. We would like to use these equations of motion to determine the control commands of each vehicle for a horizon of $H$ timesteps ahead. The cost function to be minimized during the optimization should cater to two primary objectives: One is to guide the vehicle to the target location and the other is to prevent collision with other vehicles/obstacles. 
The first component of the cost ensures that the predicted vehicle state at any step in the horizon is as close to the target state as possible. This is done by penalizing the difference between current vehicle state and target vehicle state through target cost $C_{tar}$:
\begin{equation}
\begin{split}
C_{tar} & =  \sum^{H}_{t=1} \sum^{N_{vehicle}}_{i=1} \| X^{(i)}_{t} -  X^{(i)}_{target} \|_{2} \cdot w_{pos} \\ 
& + \| \theta^{(i)}_{t} - \theta^{(i)}_{tar}|\|_{2}\cdot w_{orient} 
\end{split}
\end{equation}
For convenience  $X=[x, y]^T$ represents the position vector. $w_{pos}$ and $w_{orient}$ are the tuneable weights.

The optimization should also ensure that a vehicle does not collide with obstacles and also stays clear of other vehicles. For an obstacle with radius $r$, a penalty is introduced if a vehicle is within a margin of $r_{mar\_obs}$. The cost occurring due to collision of any vehicle with any obstacle over any of the timesteps in the horizon is described by $C_{coll\_obs}$:
\begin{equation}
\begin{split}
C_{coll\_obs} & = \sum^{H}_{t=1} \sum^{N_{vehicle}}_{i=1} \sum^{N_{obstacle}}_{j=1} [\frac{1}{\| X^{(i)}_{t} - X^{(j)} \|_{2} 
- r^{(j)}} \\ & - \frac{1}{r_{mar\_obs}} ]\cdot \Pi^{i,j}_{obs} \cdot w_{col\_obs} \\
  \Pi^{i,j}_{obs} & =
    \begin{cases}
      1 & (\| X^{(i)}_{t} - X^{(j)} \|_{2} - r^{(j)} - r_{mar\_obs}) < 0)\\
      0 & \text{otherwise}
    \end{cases}  
\end{split}
\end{equation}

Likewise, a penalty is also incurred if any vehicle $i$ collides with vehicle $j$ or is in its vicinity with a margin less than $r_{mar\_veh}$. The cost $C_{coll\_veh}$ is given as:
\begin{equation}
\begin{split}
C_{coll\_veh} & = \sum^{H}_{t=1} \sum^{N_{vehicle}-1}_{i=1} \sum^{N_{vehicle}}_{j=i+1} [\frac{1}{(\| X^{(i)}_{t} - X^{(j)}_{t}\|_{2}} \\
& - \frac{1}{r_{mar\_veh})}] \cdot  \Pi^{i,j}_{veh} \cdot w_{col\_veh} \\
  \Pi^{i,j}_{veh} & =
    \begin{cases}
      1 & (\| X^{(i)}_{t} - X^{(j)}_{t}\|_{2} - r_{mar\_veh}) < 0 \\
      0 & \text{otherwise}
    \end{cases} 
\end{split}
\end{equation}

Note that the $r_{mar\_obs}$ and $r_{mar\_veh}$ are the safety margin of the obstacles and other dynamic vehicles. When an ego-vehicle enters the safety margin of other objects, the collision cost starts to penalize it in inverse proportion to its distance to the other object. Meanwhile, $w_{col\_obs}$ and $w_{col\_veh}$ are the tuneable weights.

Finally we optimize for the control commands via:
\begin{equation}
      \min_{p,\varphi}  [C_{tar}+C_{coll\_obs}+C_{coll\_veh}]
       \label{eq:final_cost}
\end{equation}
This cost function is minimized using sequential least square programming to yield the optimal control commands.
\section{EXPERIMENTS} \label{sec:experiments}

This section provides the results of our experiments. We first give details about the data generation and model training process in Subsection \ref{subsec:data_collection}. Next, the evaluation metrics and the quantitative results are discussed in Subsection \ref{subsec:quant_results}. Lastly, the behaviour of the attention weights of our A-GNN are analyzed for a scenario with three vehicles in Subsection \ref{subsec:attention_beh}. Meanwhile, videos demonstrating the qualitative performance of our model can be visualized in the provided here: \href{https://yininghase.github.io/multi-agent-control\#Results-of-our-Model}{here}

\subsection{Data Generation and Model Training} \label{subsec:data_collection}
We are not aware of any open source platform for simultaneous control of multiple  vehicles in unconstrained environments that also considers vehicle kinematics. Therefore, we create our own as depicted in the provided codebase. The vehicles in this platform are maneuvered following the equations of motion described in Subsection \ref{subsec:label_gen}. For the purpose of training the model we collect data using this platform and determine the labels by minimizing the cost function in Equation \ref{eq:final_cost}. The data for which labels are generated contain between 1-3 vehicles and 0-4 static obstacles, for a total of around 20,961 trajectories. The start and destination states of the vehicle/obstacles are generated at random. Each trajectory is  collected for 120 timesteps. Therefore, the total number of scenarios generated are 2,515,320. Note that increasing the number of vehicles and obstacles respectively beyond 3 and 4 significantly slows down the optimization for determining the optimal control values during this data and label generation process. Nevertheless, we demonstrate in Table \ref{table:statistic of evaluation of 5 models}, that our model is still powerful enough to make inference for even more vehicles than for which it was trained. 

The collected ground truth data is split into the training and validation set with a 4:1 ratio. The data for ground truth control has a horizon of 20. However, only the control command predicted at the first step of the horizon is used to train the model. At inference time, this command is executed, which causes the vehicle to attain a new state. This new state is then fed back to the model to predict the new command. The process is iteratively repeated until the vehicle reaches the target state.  Although, the model is trained to predict only the first control command,  this does not mean that the rest of the steps in the horizon are meaningless. Rather, a longer horizon facilitates the optimization to look further into the future and take early actions to prevent a collision which would otherwise have been inevitable.
When training the model, we treat the static obstacle nodes as a special type of vehicle that has zero velocity and for which the current state is the same as the target state. The control command output by the model for such static obstacle nodes is zero for both the steering angle and pedal acceleration. Treating the obstacle nodes in this manner can help the model learn to cause the actual dynamic vehicles to not overshoot but remain stationary once they have reached their respective target state.

We use the MSE loss to train the model, which penalizes the difference between the predicted and the ground truth control variables. We use the Adam optimizer \cite{kingma2014adam} with an initial learning rate of 0.01 and weight decay of 1e-6. The learning rate is reduced by factor of 0.2 from its previous value, if the validation loss does does not reduce for 15 epochs. The number of training epochs is set to be 500 but training is prematurely stopped, if there is no decrease in the validation loss for 50 epochs.

\subsection{Quantitative results}\label{subsec:quant_results}

\begin{table*}[h]
\centering
\resizebox{\textwidth}{!}{
 \begin{tabular}{||c c c c c c c c c c c c c||} 
 \hline
  &  & \multicolumn{5}{|c|}{success to goal rate 	$\uparrow$} & \multicolumn{5}{|c|}{collision rate  $\downarrow$} & step efficiency $\uparrow$ \\ 
 \hline
 Number of.& Number of.& Our & GAIN  & TransformerConv  & EdgeConv & No  & Our & GAIN & TransformerConv & EdgeConv  & No  & Our\\ [0.5ex] 
  Vehicles.& Obstacles.& Model & \cite{9827155}  & \cite{shi2020masked}  & \cite{yue2019dynamic} & Graph & Model & \cite{9827155} & \cite{shi2020masked}  & \cite{yue2019dynamic} & Graph & Model\\ [0.5ex] 
 \hline\hline
1 & 0 & \textbf{1.0000} & 0.9850 & 0.9894 & 0.1890 & \textbf{1.0000} & - & - & - & - & - & 1.0000\\
\hline
1 & 1 & \textbf{0.9677} & 0.7792 & 0.7580 & 0.7610 & 0.8124 & \textbf{4.9616E-05} & 5.6631E-03 & 5.3226E-03 & 5.4148E-03 & 5.5137E-03 & 1.0000 \\
\hline
1 & 2 & \textbf{0.8991} & 0.5751 & 0.5721 & 0.5771 & 0.6593 & \textbf{3.3231E-04} & 1.1459E-02 & 1.1451E-02 & 1.1487E-02 & 1.1607E-02 & 1.0000 \\
\hline
1 & 3 & \textbf{0.8127} & 0.4707 & 0.4929 & 0.4902 & 0.5682 & \textbf{6.6306E-04} & 1.5154E-02 & 1.6024E-02 & 1.6851E-02 & 1.6859E-02 & 1.0000 \\
\hline
1 & 4 & \textbf{0.7361} & 0.3582 & 0.3922 & 0.3959 & 0.3582 & \textbf{1.2774E-03} & 1.9003E-02 & 2.1938E-02 & 2.2296E-02 & 2.2594E-02 & 1.0000 \\
\hline
2 & 0 & \textbf{0.9984} & 0.6285 & 0.6600 & 0.6436 & 0.5881 & \textbf{1.1006E-05} & 8.6178E-03 & 7.6941E-03 & 8.0362E-03 & 9.7184E-03 & 2.0239 \\
\hline
2 & 1 & \textbf{0.9634} & 0.6251 & 0.6219 & 0.6120 & 0.5627 & \textbf{1.6017E-04} & 1.0111E-02 & 9.6096E-03 & 1.0098E-02 & 1.2165E-02 & 1.9831 \\
\hline
2 & 2 & \textbf{0.8676} & 0.5538 & 0.5624 & 0.5665 & 0.5379 & \textbf{4.5357E-04} & 1.2926E-02 & 1.2796E-02 & 1.3069E-02 & 1.4399E-02 & 1.9396 \\
\hline
2 & 3 & \textbf{0.7792} & 0.4838 & 0.5209 & 0.5156 & 0.5049 & \textbf{6.7795E-04} & 1.6143E-02 & 1.6039E-02 & 1.6378E-02 & 1.7036E-02 & 1.9582 \\
\hline
2 & 4 & \textbf{0.6781} & 0.4439 & 0.4982 & 0.4862 & 0.4913 & \textbf{1.1135E-03} & 1.8801E-02 & 1.8711E-02 & 1.9202E-02 & 1.8788E-02 & 1.8513 \\
\hline
3 & 0 & \textbf{0.9943} & 0.4227 & 0.4339 & 0.4401 & 0.3993 & \textbf{8.8107E-05} & 1.4238E-02 & 1.4047E-02 & 1.3778E-02 & 1.6330E-02 & 2.9871 \\
\hline
3* & 1* & \textbf{0.9706} & 0.4149 & 0.4193 & 0.4279 & 0.3708 & \textbf{3.0382E-04} & 1.5672E-02 & 1.5655E-02 & 1.4713E-02 & 1.8927E-02 & 2.9334 \\
\hline
3* & 2* & \textbf{0.9302} & 0.4102 & 0.4174 & 0.4126 & 0.3582 & \textbf{5.9881E-04} & 1.7267E-02 & 1.7393E-02 & 1.6426E-02 & 2.0821E-02 & 2.9399 \\
\hline
3* & 3* & \textbf{0.8903} & 0.4023 & 0.4023 & 0.3938 & 0.3499 & \textbf{8.9677E-04} & 1.8393E-02 & 1.9646E-02 & 1.8697E-02 & 2.2789E-02 & 2.8887 \\
\hline
3* & 4* & \textbf{0.8328} & 0.3818 & 0.3834 & 0.3716 & 0.3328 & \textbf{1.6090E-03} & 2.0214E-02 & 2.2045E-02 & 2.1205E-02 & 2.5397E-02 & 3.0669 \\
\hline
4* & 0* & \textbf{0.9807} & 0.2850 & 0.2895 & 0.3108 & 0.2614 & \textbf{2.7650E-04} & 1.8619E-02 & 1.9967E-02 & 1.8791E-02 & 2.2929E-02 & 3.9478 \\
\hline
4* & 1* & \textbf{0.9550} & 0.3088 & 0.3048 & 0.3187 & 0.2526 & \textbf{5.7179E-04} & 1.8783E-02 & 2.0446E-02 & 1.8025E-02 & 2.4251E-02 & 4.0105 \\
\hline
4* & 2* & \textbf{0.9279} & 0.3031 & 0.2905 & 0.2970 & 0.2379 & \textbf{9.4375E-04} & 1.9565E-02 & 2.1960E-02 & 1.9552E-02 & 2.5899E-02 & 3.9228 \\
\hline
4* & 3* & \textbf{0.8853} & 0.3077 & 0.2864 & 0.2930 & 0.2344 & \textbf{1.4612E-03} & 1.9224E-02 & 2.3747E-02 & 2.1138E-02 & 2.7634E-02 & 3.7834 \\
\hline
5* & 0* & \textbf{0.9590} & 0.2243 & 0.1973 & 0.2208 & 0.1797 & \textbf{5.8217E-04} & 2.0827E-02 & 2.5964E-02 & 2.3437E-02 & 2.9177E-02 & 4.8234 \\
\hline
5* & 1* & \textbf{0.9285} & 0.2361 & 0.2078 & 0.2306 & 0.1699 & \textbf{1.0483E-03} & 2.0363E-02 & 2.6115E-02 & 2.1317E-02 & 3.0344E-02 & 4.9839 \\
\hline
5* & 2* & \textbf{0.9037} & 0.2559 & 0.2125 & 0.2254 & 0.1663 & \textbf{1.4063E-03} & 1.8332E-02 & 2.6607E-02 & 2.2013E-02 & 3.1475E-02 & 4.2917 \\
\hline
6* & 0* & \textbf{0.9209} & 0.1967 & 0.1347 & 0.1631 & 0.1207 & \textbf{1.1376E-03} & 1.9465E-02 & 3.1612E-02 & 2.7085E-02 & 3.5283E-02 & 7.0979 \\
\hline
6* & 1* & \textbf{0.8949} & 0.1947 & 0.1556 & 0.1744 & 0.1195 & \textbf{1.5096E-03} & 1.7424E-02 & 3.0917E-02 & 2.4239E-02 & 3.5972E-02 & 5.5704 \\
\hline
6* & 2* & \textbf{0.8717} & 0.1697 & 0.1479 & 0.1660 & 0.1152 & \textbf{1.8932E-03} & 1.5176E-02 & 3.1828E-02 & 2.4354E-02 & 3.6918E-02 & 2.8350 \\
\hline
\end{tabular}
}
\caption{Shows the performance of the five different models i.e. our Model, GAIN \cite{9827155}, TransformerConv \cite{shi2020masked}, EdgeConv \cite{yue2019dynamic} and No Graph (MLP) when measured against the \emph{success-to-goal rate} and \emph{collision rate} metrics. The evaluation is done on completely unseen test data comprising of scenarios with 1-6 vehicles and 0-4 obstacles as shown by the corresponding rows. Each row in the table is evaluated on 4062 scenarios. The rows labeled with an asterisk (*) are those vehicle-obstacle combinations that were not even in the training set. The training set only comprised of 1-3 and 0-4 obstacles. Therefore, these rows are particularly interesting in demonstrating that our model has generalization capability giving good performance even in these situations. Being the best performing, the \emph{step efficiency} metric is also given for our method in the last column. It shows that our method is generally more efficient than running the vehicles one by one.}
\label{table:statistic of evaluation of 5 models}
\end{table*}

We evaluate the performance of our method in an online setting on a randomly generated dataset not seen during training. The test set contains scenarios having  1-6 vehicles and 0-4 obstacles. The model needs to predict the appropriate control commands such that all the vehicles in the test scenarios reach their target state without colliding with others. The online performance of the model is evaluated against three criterion: \\

\noindent{\textbf{1) Success-to-goal rate:}} is defined by the percentage of vehicles that successfully reach their target state within a tolerance without colliding with other objects along the way. The location tolerance is set to be 1.25 meters from the vehicle centers and angle tolerance is set to 0.2 radians. Higher value for this metric is better. \\

\noindent{\textbf{2) Collision rate:}} is defined as the total number of collisions caused by a model divided by the total distance travelled by all vehicles. Note that this metric is meaningless when evaluation is being done for one vehicle with no obstacles. This is because no collisions are expected to occur in such a scenario. A lower value of this metric is better. Lastly, note that the inverse of this metric describes the average total distance travelled before a collision happens. \\

\noindent{\textbf{3) Step Efficiency:}}
A naive approach to solve the problem of navigating all the vehicles to the goal while avoiding collision is to control the individual vehicles one by one while keeping other vehicles stationary. This considerably simplifies the problem and is akin to having one dynamic agent with multiple static obstacles. However, this approach is inefficient taking more steps to solve the whole problem with many vehicles in the scene. Therefore, we introduce the step efficiency metric, to show the advantage of our model on tackling all the vehicles simultaneously. It is the ratio of the lower bound of the number of steps required by running the vehicles one by one with other vehicle kept stationary divided by the number of steps used by navigating all the vehicles simultaneously using our approach. A higher value for this metric is better. 

For the step efficiency metric, calculating the total number of steps by running the the vehicles one after the other is not trivial. This is because the total number of steps is influenced by the order in which the vehicles are run. Therefore, to ensure that the the step efficiency metric is not affected by the permutation order, we use the lower bound of the actual number of steps for running the vehicles one after the other. To do this, all other vehicles are ignored when one of the vehicles is run with its steps being counted. This leads to a simplified version of the original task since the vehicle being run only has to consider the presence of static obstacles. This leads to either an equal or lesser number steps taken to reach its destination. The same is done for all the other vehicles and their steps are summed to yield the lower bound of the actual number of steps taken. Since the summation is a permutation invariant function, the step efficiency metric is therefore not influenced by the vehicle order. This is implemented by simply removing the edges between vehicles in our GNN model so that all the vehicles run simultaneously rather than one-by-one to get the total steps.\\

\noindent{\textbf{Comparision with other methods:}}
We make comparison with 3 additional Graphical Neural Network architectures: namely the attention based GAIN \cite{9827155} and TransformerConv \cite{shi2020masked} and the non-attention based EdgeConv \cite{yue2019dynamic}. Comparision is also made with the naive Multi-Layer Perceptron (MLP) model which has a similar architecture to our model but does not utilize any graphical edges in its structure. \\

\noindent{\textbf{Training:}} The training process for all these models is similar as our method described in Subsection \ref{subsec:data_collection}. The difference is in the respective architectures. GAIN \cite{9827155} and TransformerConv \cite{shi2020masked} do not utilize the relative information of the neighbouring edge nodes as was done for our case in Equation \ref{eq:Fatt}. They also do not utilize a joint trainable U-Net inspired architecture with shared encoder weights for the attention mechanism.  Meanwhile, EdgeConv \cite{yue2019dynamic} is different from our architecture in that a node does not attend to other nodes. Meanwhile, the MLP does not use any graphical edges in its structure. Note that GAIN \cite{9827155}, had utilized the historical trajectory information of the road vehicles as input. We are only using the current and target state and therefore do not require processing of any temporal information. Moreover, rather than having the entire road network in the map topology guiding the vehicles, our environment is unconstrained. Therefore, to adapt the GAIN into our pipeline, we replace all our graph attention layers with the graph attention isomorphism operator introduced in \cite{9827155} without needing any historical trajectory information and the entire road map topology. For TransformerConv \cite{shi2020masked} and EdgeConv \cite{yue2019dynamic} we directly use the implementation from \cite{Fey/Lenssen/2019}. \\ 

\noindent{\textbf{Evaluation:}} Experimental results for all the models are shown in Table \ref{table:statistic of evaluation of 5 models}. Note that our approach consistently performs better than all the other approaches. The model even performs well when there are 4-6 vehicles in the scene.  It is important to highlight here that our training data only contained up to a maximum of 3 vehicles. This demonstrates the predictive power of our A-GNN which has generalized to work on even more number of vehicles than it was trained. 

\begin{figure*}[ht]
\centering
    \subfigure[\centering
    {Our Model}]{{\includegraphics[width=0.19\textwidth]{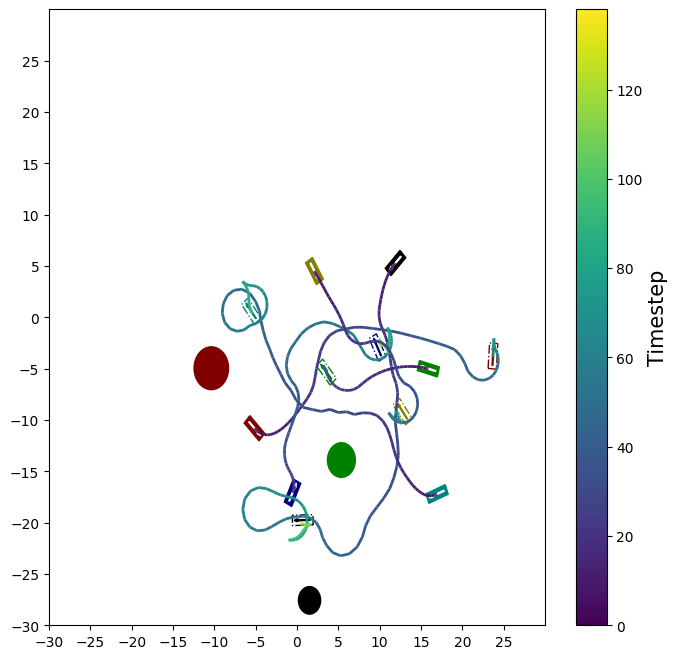}}}
    \subfigure[\centering
    {GAIN}]{{\includegraphics[width=0.19\textwidth]{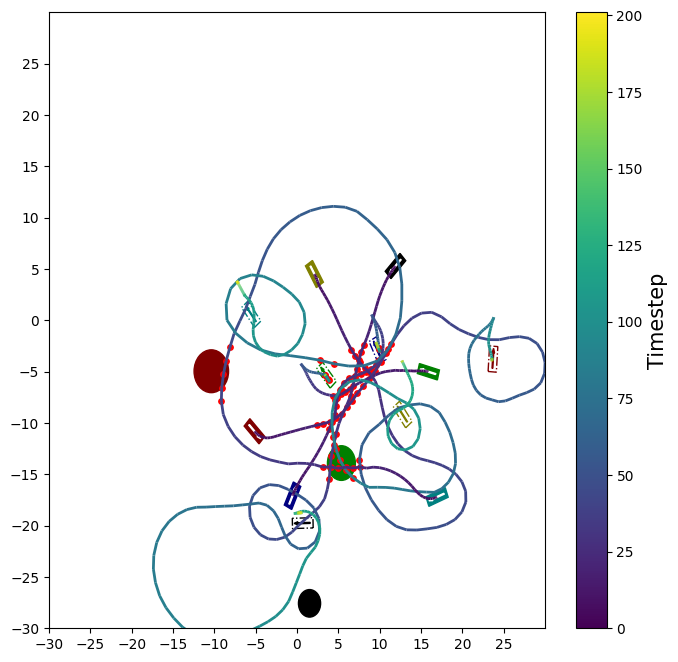}}}
    \subfigure[\centering
    {TransformerConv}]{{\includegraphics[width=0.19\textwidth]{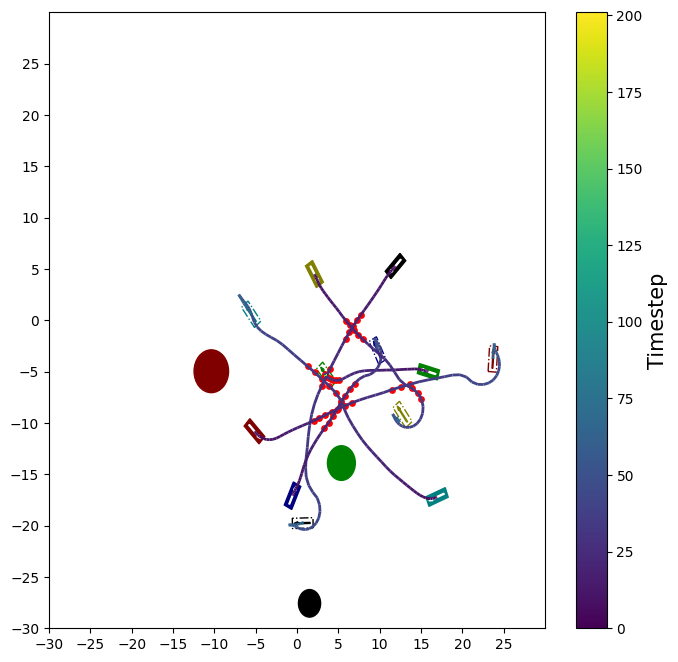}}}
    \subfigure[\centering
    {EdgeConv}]{{\includegraphics[width=0.19\textwidth]{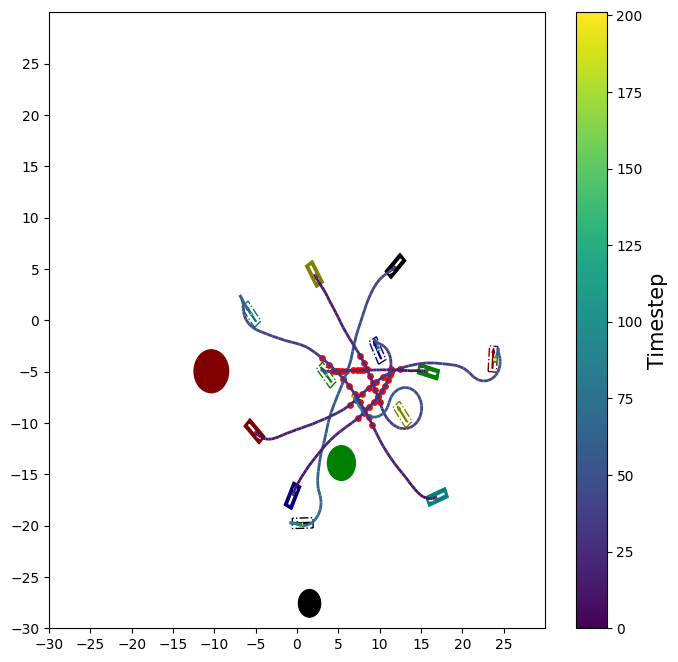}}}
    \subfigure[\centering
    {No Graph}]{{\includegraphics[width=0.19\textwidth]{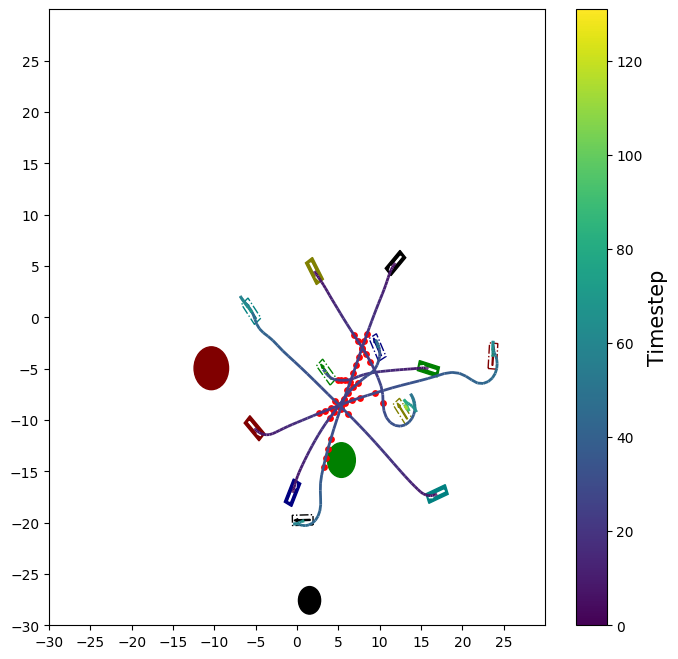}}}
    \quad
    \caption{Shows the trajectory traversed by the vehicles as a result of applying the control commands predicted by the five different models for a scenario containing 6 vehicles and 3 obstacles. The red dots on the trajectories show the point of collision between the vehicles. Except for our model, all other models have plenty of collisions. Video can be found: \href{https://yininghase.github.io/multi-agent-control\#Comparison-with-Other-Models}{here}.}
    \label{fig:trajectories_compare_of_5_models}
\end{figure*}

We now give some observations about the performance of our A-GNN in Table \ref{table:statistic of evaluation of 5 models} which is superior to the other three Graphical Neural Network architectures and the naive MLP model without edges. It can be seen that the collision rate of our model is much lower and the success-to-goal rate is also higher than all the others. The reason for poor performance of the other three models is that they only seem to learn to reach the target but do not learn how to avoid collision. Figure \ref{fig:trajectories_compare_of_5_models} shows that in pursuit of reaching their desired destination, the vehicles controlled by the other models collide with one another or with other obstacles. Hence, they have an extremely high collision rate. The reason why our model outperforms all the other models is because of the difference in architectures. Primarily we are using a U-net inspired attention mechanism with shared encoder weights along with concatenating relative information between the self-node and the neighbouring nodes when determining the attention weights. \ref{subsec:model-ablation} shows the importance of contribution of these components of the architecture. Removal of either the U-Net architecture or not concatenating the relative information reduces the performance.

Furthermore, Figure \ref{fig:trajectories_compare_of_5_models} shows our model can generalize when there are 6 vehicles in the environment. Such a scenario was not present in the training dataset. The training dataset was limited to a maximum of 3 vehicles. Meanwhile, the other models show poor performance on these scenarios with plenty of collisions among the vehicles. 

Note that the in Table \ref{table:statistic of evaluation of 5 models} of the main paper, the success-to-goal rate for our model is not a perfect score of 1. The reason the model fails to always reach the target destination is that it tends to behave conservatively. Rather than taking the risk of collision, it sometimes stops mid-way before other objects to avoid collision. This is because the ground truth data obtained from the optimization is not always prefect. Nevertheless, it is important to keep such failed samples in the dataset for training as they teach the model to learn to stop before other objects to avoid collision. If these failure samples are removed from the training set, then the model performs worse as it starts colliding with other obstacles.

\begin{figure}[ht]
\centering
\includegraphics[width=0.9\columnwidth]{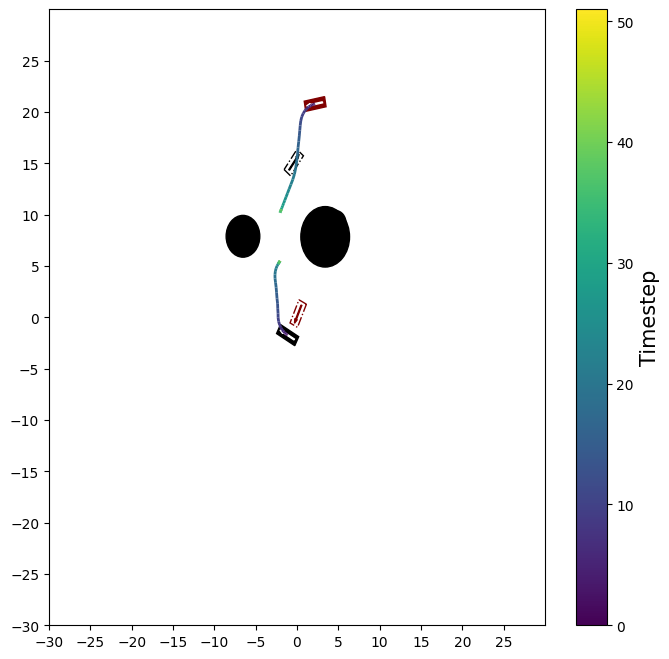}
\caption{An example of a conservative optimization yielding a sample trajectory wherein the vehicles halt rather than take the risk of collision in an attempt to reach their  destination by trying to pass between the two obstacles. }
\label{fig:failed_opt_trajectory}
\end{figure}

Note that the success-to-goal rate for our model is not a perfect score of 1. The reason the model fails to always reach the target destination is because it tends to behave conservatively. Rather than taking the risk of collision, it sometimes stops mid-way before other objects to avoid collision. This is because the ground truth data obtained from the optimization is not always prefect. Figure \ref{fig:failed_opt_trajectory} shows an example of such conservative behaviour wherein the optimization failed to produce the requisite control commands to reach the target. Rather, the vehicle gets stuck midway. Nevertheless, it is important to keep such failed samples in the dataset for training as they teach the model to learn to stop before other objects to avoid collision. If these failure samples are removed from the training set, then the model performs worse as it starts colliding with other obstacles.

Lastly, since our model is the best performing among all others, the last column in Table \ref{table:statistic of evaluation of 5 models} also provides the \emph{step efficiency} metric for it. It can be observed that our method of executing the control commands for all vehicles simultaneously tends to be faster than running the vehicles one by one. 

\subsection{Attention Weights Analysis}\label{subsec:attention_beh}

\begin{figure*}[ht]
\centering
    \subfigure[\centering
    {$t_{1}$}]{{\includegraphics[width=\columnwidth]{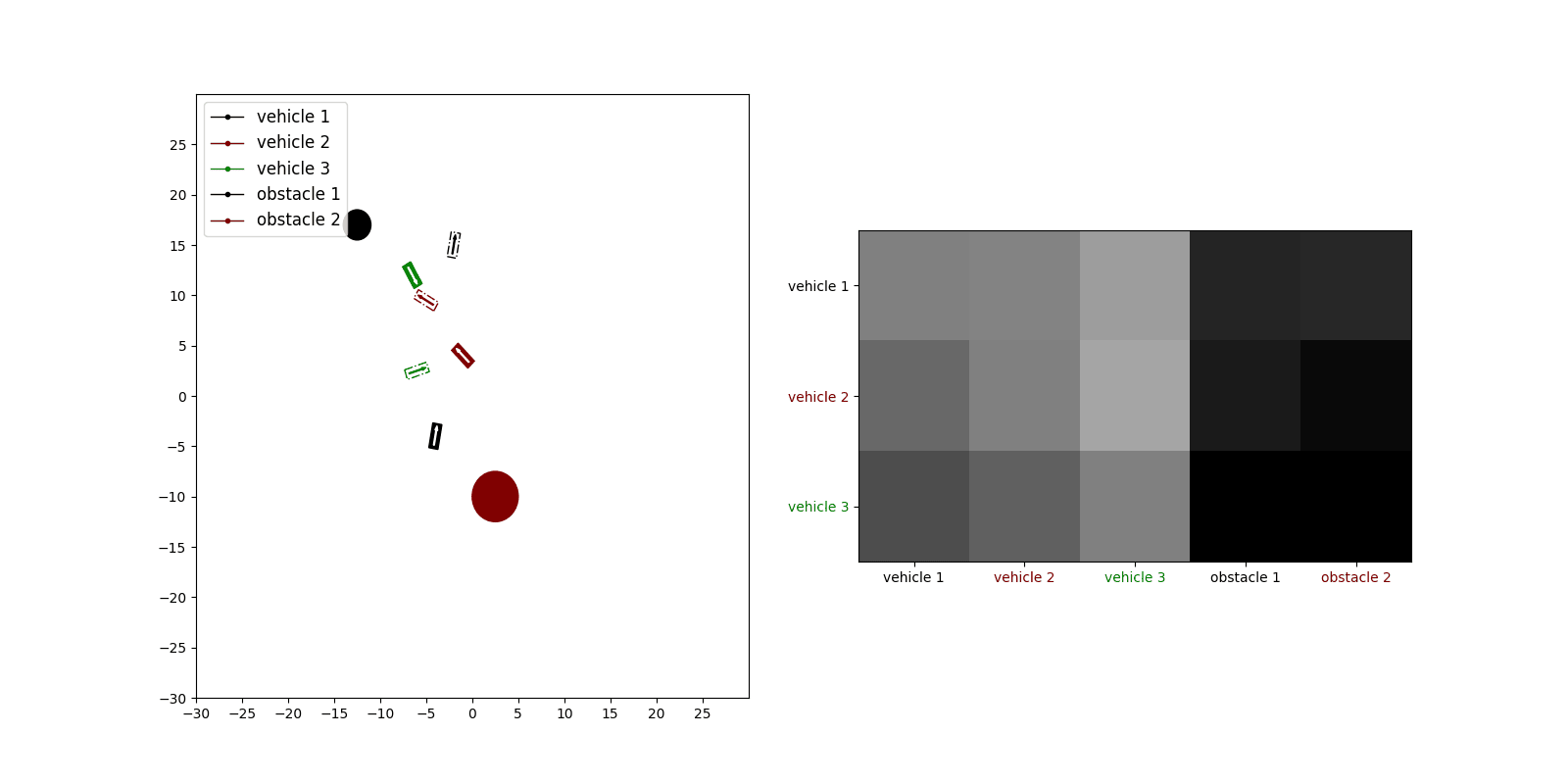}}}
    \subfigure[\centering
    {$t_{2}$}]{{\includegraphics[width=\columnwidth]{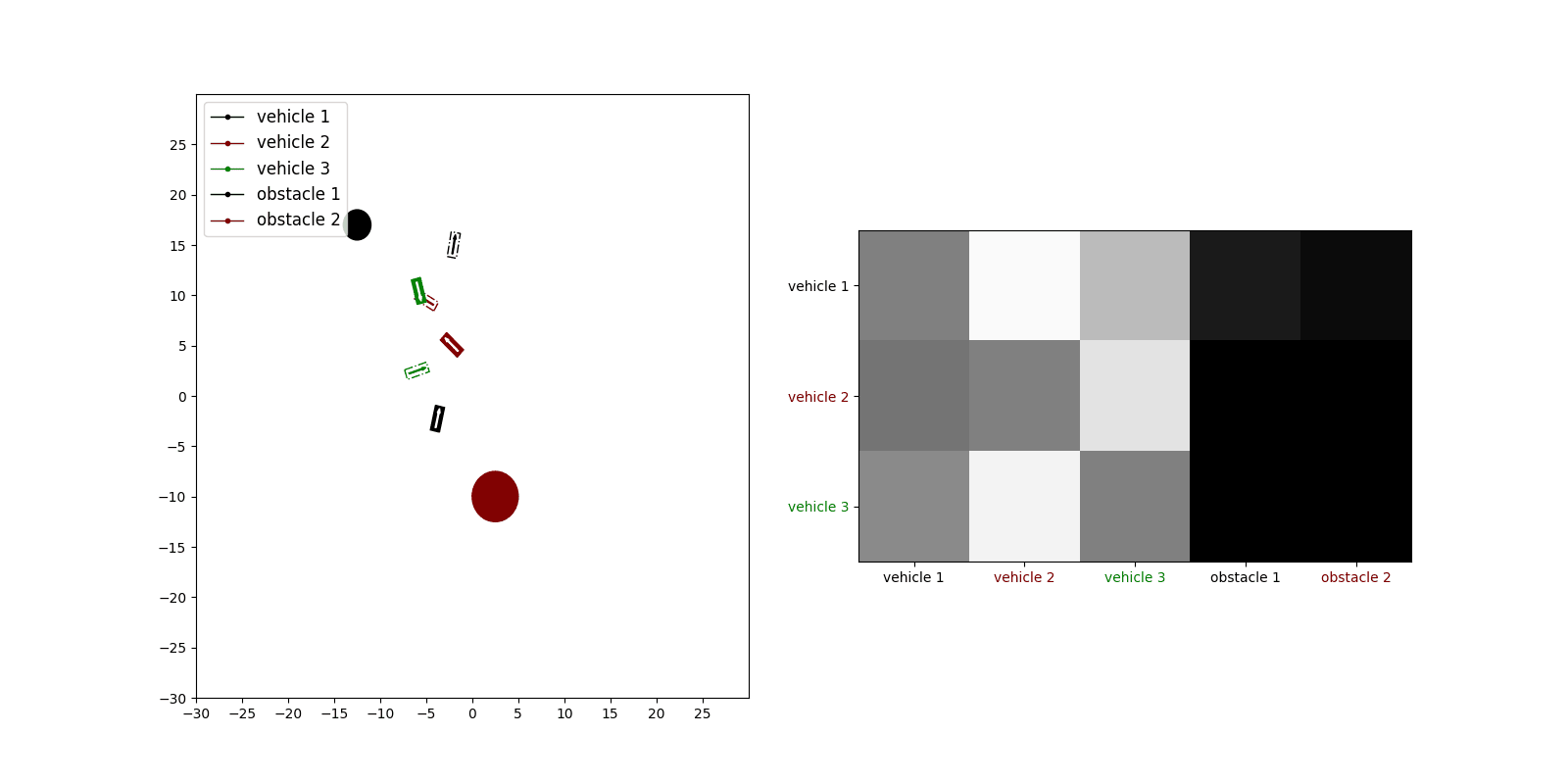}}}
    \quad
    \subfigure[\centering
    {$t_{3}$}]{{\includegraphics[width=\columnwidth]{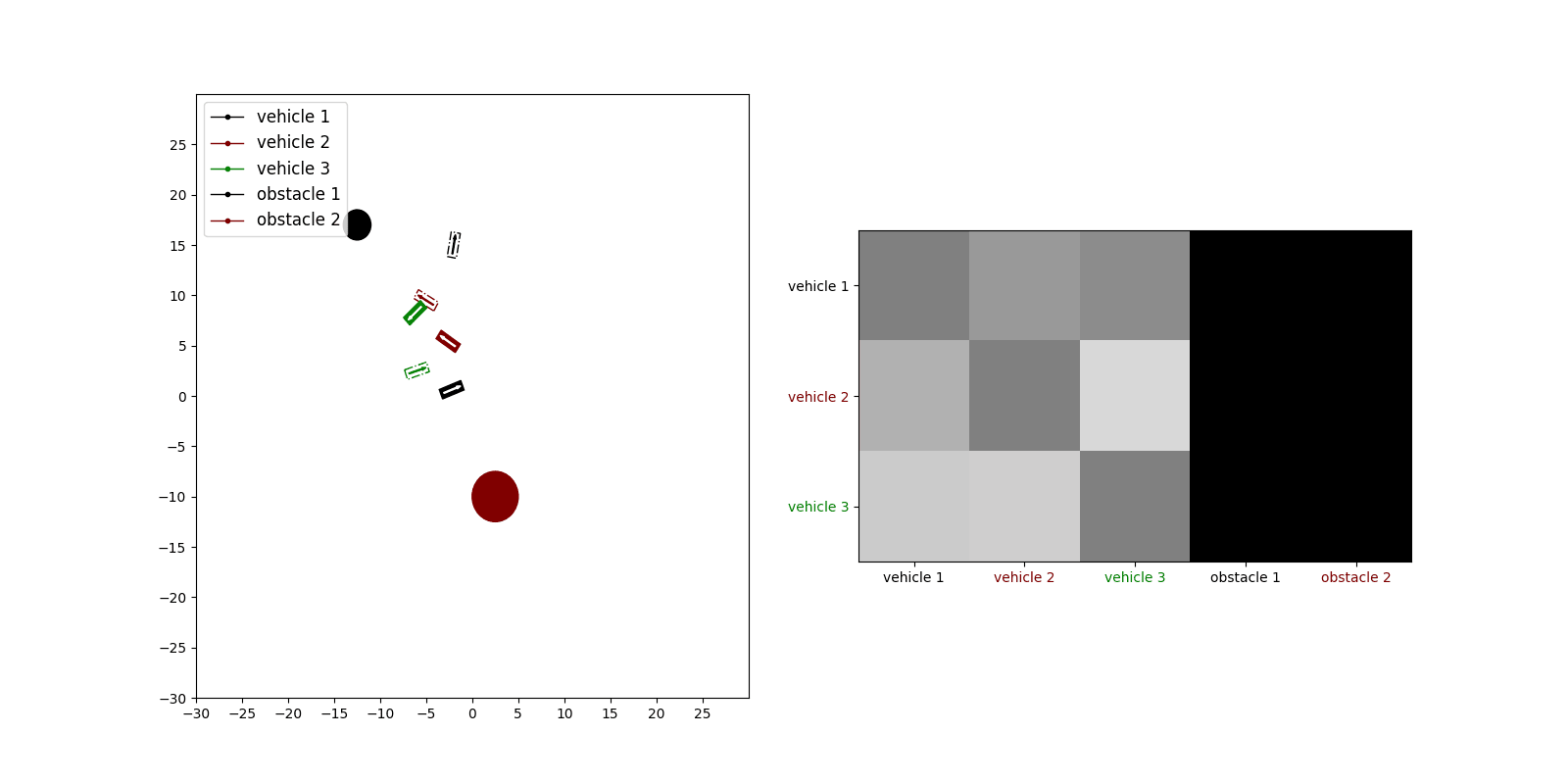}}}
    \subfigure[\centering
    {$t_{4}$}]{{\includegraphics[width=\columnwidth]{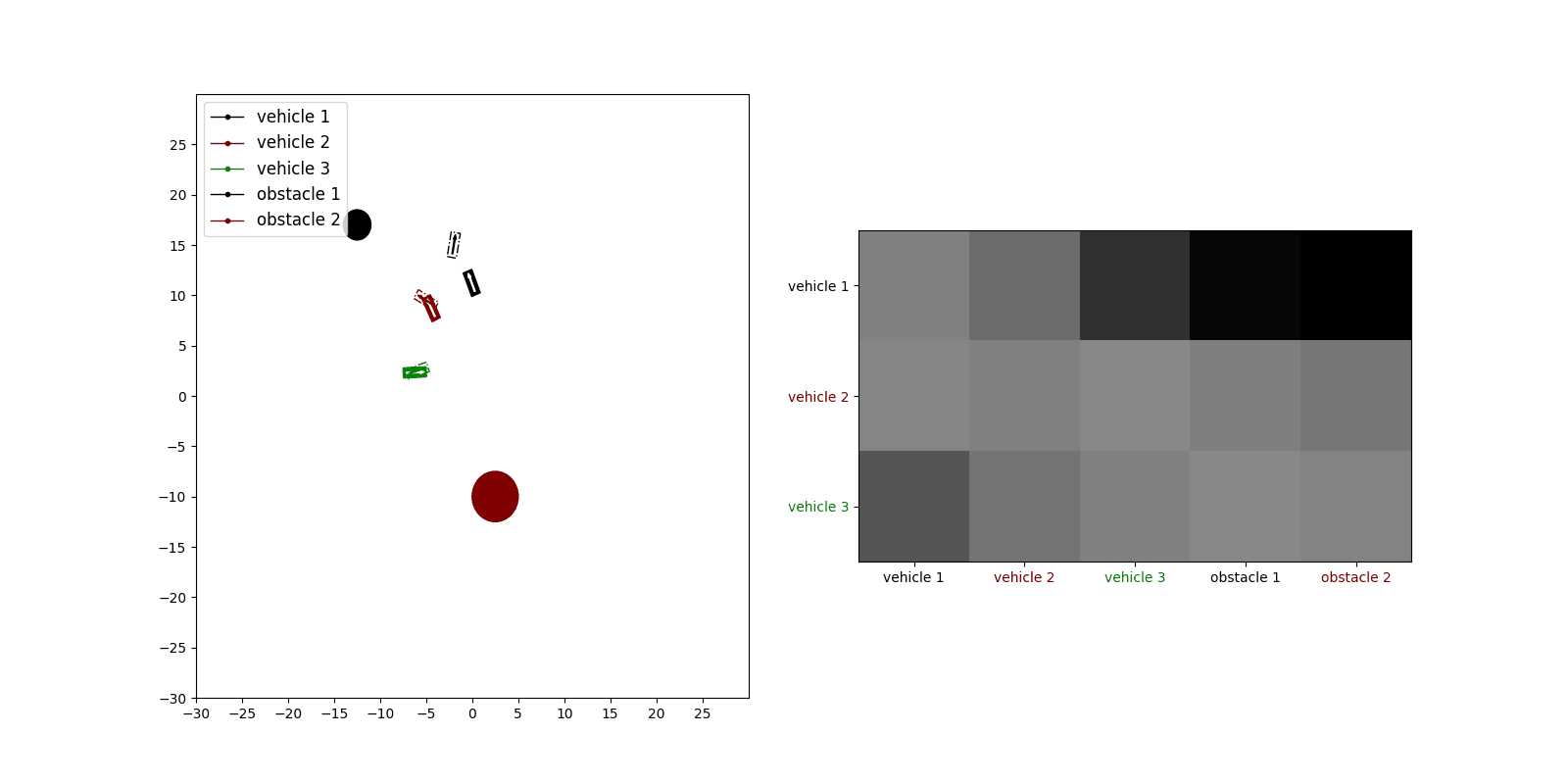}}}
    \caption{\textbf{Predicted Trajectories of our model with Attention Logits}: The color of pixel at row $i$ column $j$ shows the attention logits of object $i$ on object $j$. The lower the logits value is, the darker is the color and vice versa. The gray value on the diagonal means that a vehicle is neutral to attending to itself. Corresponding video can be found in the codebase: \href{https://yininghase.github.io/multi-agent-control\#Attention-Mechanism-of-our-Model}{here}}
    \label{fig:trajectories_of_our_model_with_attention_logits}
\end{figure*}

To get some intuition about the behavior of our model, we visualize the mean of attention logits from all the graph attention layers of our model. Figure \ref{fig:trajectories_of_our_model_with_attention_logits} shows a visualization of these attention logits as a matrix for 4 timesteps $t_1$, $t_2$, $t_3$ and $t_4$ of a scenario with 3 vehicles and 2 obstacles. The rows in the attention matrix correspond to the vehicle of interest. The columns show which vehicle/obstacle is being attended to. A lighter shade in the attention matrix depicts high attention and a darker shade represents lack of attention. Because the static obstacles are stationary and are never a hindrance for any of the 3 vehicles, their attention logits of all the vehicles on them is very low (dark pixel on the last 2 columns) for all the 4 timesteps. At time $t_1$, the vehicles are relatively far away from each other, so the attention logits are relatively low. At $t=t_{2}$, the \textbf{\textcolor{red}{red}} vehicle \textcolor{black}{(vehicle 2)} is in the way of the \textbf{black} vehicle (vehicle 1), the \textbf{\textcolor{green}{green}} vehicle \textcolor{black}{(vehicle 3)} is in the way of \textbf{\textcolor{red}{red}} vehicle \textcolor{black}{(vehicle 2)}, the \textbf{\textcolor{red}{red}} vehicle \textcolor{black}{(vehicle 2)} is in the way of \textbf{\textcolor{green}{green}} vehicle \textcolor{black}{(vehicle 3)} and all the 3 vehicles are close to each other. So each vehicle pays more attention to the other vehicles that are on their way to the target and in close proximity. These 3 corresponding pixels are therefore very bright. At $t=t_{3}$, the \textbf{\textcolor{red}{red}} vehicle \textcolor{black}{(vehicle 2)} decides to stop and let the \textbf{\textcolor{green}{green}} vehicle \textcolor{black}{(vehicle 3)} pass through. So the attention of the \textbf{\textcolor{green}{green}} vehicle \textcolor{black}{(vehicle 3)} on the \textbf{\textcolor{red}{red}} vehicle \textcolor{black}{(vehicle 2)} goes down. However the \textbf{\textcolor{red}{red}} vehicle \textcolor{black}{(vehicle 2)} needs to wait until the \textbf{\textcolor{green}{green}} vehicle \textcolor{black}{(vehicle 3)} passes through. So the attention of \textbf{\textcolor{red}{red}} vehicle \textcolor{black}{(vehicle 2)} on the \textbf{\textcolor{green}{green}} vehicle \textcolor{black}{(vehicle 3)} is still high. And the \textbf{black} vehicle (vehicle 1) has found its way to go around the \textbf{\textcolor{red}{red}} vehicle \textcolor{black}{(vehicle 2)}, so the attention of the \textbf{black} vehicle (vehicle 1) on the \textbf{\textcolor{red}{red}} vehicle \textcolor{black}{(vehicle 2)} also goes down. Finally, all the vehicles reach their goal, and their attentions on each other reduce.
This trivial qualitative example demonstrates, that our model is able to learn driving behaviour analogous to how a human would make decisions. However, more experiments and quantitative studies would be required to better understand the reasoning process behind these graphical neural architectures. Therefore, this work can potentially be used as a starting point for interpreting the decision making capabilities of trained models before their deployment in relevant and related applications.

\section{Future Work}\label{sec:future_work}

As we discussed in Section \ref{sec:related_work}, concurrent Reinforcement Learning (RL) methods do not consider the vehicle kinematics and also tend to be heavily sample-inefficient requiring far more steps than our method to train the model. However, the advantage is that they can be trained without labels obtained from optimization. Therefore, to exploit this aspect, we can extend this work by using our pretrained model as an initializer to prevent the cold-start problem associated with Reinforcement Learning. This way the model can be trained by progressively adding more vehicles into the environment. This prevents the cumbersome offline optimization process.

\section{Conclusion}\label{sec:conclusion}

In this paper, we proposed an attention based graphical neural network model. It is capable of predicting the control commands for multiple agents for reaching their desired destination without collision. We demonstrated that utilizing information of both the ego-vehicle and neighboring agents in the graphical layers helps the model generalize well. In fact, the model performs even on situations with more vehicles and obstacles than those in the training set. \\

\textbf{\textit{Acknowledgements:}} We thank Marc Brede for helping with the initial setup of the label generation process in the early phase of the project. \\

\bibliographystyle{IEEEtran}
\bibliography{IEEEabrv, root}
\clearpage
\section{Supplementary}\label{sec:supplementary}
\subsection{Details of the U-Net inspired attention mechanism}\label{subsec: U-Attention origin}

Recall that in Section \ref{sec:framework} of the main paper, we had used joint trainable parameters of a U-Net inspired architecture with shared encoder weights for determining $F_{value}, F_{query}, F_{key}$. In this subsection we elaborate more on this architecture and its implementation.

The idea of U-Attention originates from first integrating together the characteristics of TransformerConv\cite{shi2020masked} and EdgeConv\cite{yue2019dynamic}. TransformerConv applies attention on the neighbouring nodes but does not concatenate information of the ego-node.  In contrast, EdgeConv concatenates relative information of the neighbouring nodes with that of the ego-node. However, information aggregation is done without deciding which node to pay more attention to. Our first step is to first combine these two ideas, i.e.:
\begin{equation}\label{eq:Fatt_linear}
  F^{l+1}_{Att}(z^{l}_{i},   z^{l}_{j}) = W^{l}_{self} \cdot z^{l}_{i} + \sum_{j \in N_{i}} \alpha_{ij} \cdot W_{value}(z^{l}_{i} | z^{l}_{i}-z^{l}_{j})  
\end{equation}
where $\alpha_{ij} = {{W_{query}(z^{l}_{i})^{T} \cdot W_{key}(z^{l}_{j})} \over { \sum_{k \in N_{i}} W_{query}(z^{l}_{i})^{T} \cdot W_{key}(z^{l}_{k})}}$ and $W_{value}, W_{query}, W_{key}$ are the trainable parameter matrices. For simplicity, we set $x^{l}_{j} = (z^{l}_{i} | z^{l}_{i}-z^{l}_{j})$. If we only consider operations within a layer, we can eliminate the need for the layer index $l$, so $x^{l}_{j}$ can be further simplified to $x_{j}$. The key, query and value of $x_{j}$ is respectively denoted by $K_{j}$, $Q_{j}$ and $V_{j}$. In the traditional transformer block, $K_{j}$, $Q_{j}$ and $V_{j}$ can be calculated by $K_{j} = W_{key} \cdot x_{j}$, $Q_{j} = W_{query} \cdot x_{j}$ and $V_{j} = W_{value} \cdot x_{j}$.

Next step, we look at the formulation of self-attention weights $\alpha_{ij}$:
\begin{equation}
\begin{split}
    \alpha_{ij} & = {{K_{j}^T \cdot Q_{j}} \over { \sum_{k \in N_{i}} K_{k}^T \cdot Q_{j}}} \\
                & = {{(W_{key} \cdot x_{j})^T \cdot (W_{query} \cdot x_{j})} \over { \sum_{k \in N_{i}} (W_{key} \cdot x_{k})^T \cdot (W_{query} \cdot x_{j})}} \\
                & = {{ x_{j}^T \cdot (W_{key}^T \cdot W_{query}) \cdot x_{j}} \over { \sum_{k \in N_{i}}  x_{k}^T \cdot (W_{key}^T \cdot W_{query}) \cdot x_{j}}}
                \label{eq:matrices}
\end{split}
\end{equation}

Note that we can treat the matrix multiplication ($W_{key}^{T} \cdot W_{query}$) in the above equation as a single entity, in which the output has the same dimension as the input just as in the case of an autoencoder.  Here we can treat $W_{key}$ as the encoder while $W_{query}$ as the decoder. Both the encoder and the decoder have only one single linear layer without any non-linearity. Inspired by this interpretation, we use the auto-encoder to directly calculate attention $\alpha$. Skip connections are added between the encoder and decoder for stability. With these changes we have the neural networks $F_{query}, F_{key}$ instead of the original matrices $W_{key}$ and $W_{query}$ and Equation \ref{eq:matrices} is modified to:
\begin{equation}\label{eq:new_alpha}
\begin{split}
\alpha_{ij} & = {{F_{query}(F_{key}(x_{j}))^T \cdot x_{j}} \over { \sum_{k \in N_{i}} F_{query}(F_{key}(x_{k}))^T \cdot x_{j}}} \\
            & = {{F_{query\_key}(x_{j}))^T \cdot x_{j}} \over { \sum_{k \in N_{i}} F_{query\_key}(x_{k}))^T \cdot x_{j}}} \\
\end{split}
\end{equation}

Meanwhile, the second term of Equation \ref{eq:Fatt_linear} can be re-written as:
\begin{equation}
\begin{split}
\sum_{j \in N_{i}} \alpha_{ij} \cdot W_{value}(z_{i} | z_{i}-z_{j}) & = \sum_{j \in N_{i}} \alpha_{ij} \cdot W_{value} \cdot x_{j}  \\ & = \sum_{j \in N_{i}} \alpha_{ij} \cdot V_{j}
\end{split}
\end{equation}
where $\alpha_{ij}$ is calculated by a U-net attention network described earlier. It is easy to notice that this U-net has a latent vector $F_{key}(x_j)$ at the output of the encoder, which already encodes information about $x_j$. This information can be used to determine the value $V$ as it is also a function of $x_j$. Originally, the value $V$ is given by a simple matrix multiplication between the input $x_j$ and the matrix $W_{value}$. Instead of this we pass the the latent vector $F_{key}(x)$ through an independent decoder $F_{value}$ to determine the value of $V$. Skip connections are additionally added from encoder $F_{key}$ to this decoder head $F_{value}$. So the value can be calculated by:
\begin{equation}\label{eq:new_value}
\begin{split}
V_{i} & = F_{value}(F_{key}(x_{j})) \\
      & = F_{value\_key}(x_{j})
\end{split}
\end{equation}

\begin{figure*}[!h]
\centering
\includegraphics[width=\textwidth]{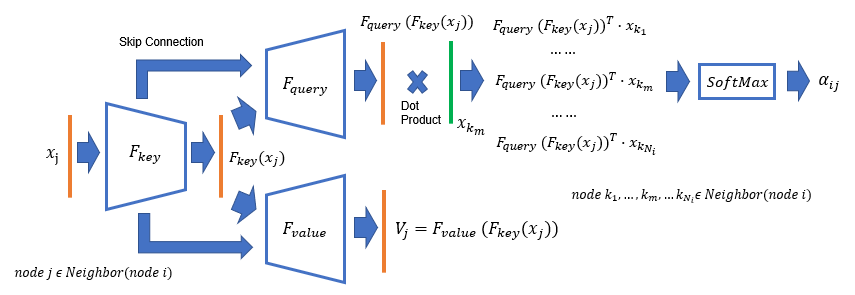}
\caption{\label{fig:U-Attention Block} \textbf{Schematics of the U-Attention Block}}
\end{figure*}

Hence, by replacing the linear layers $W_{key}$, $W_{query}$ and $W_{value}$ in Equation \ref{eq:Fatt_linear} with Equation \ref{eq:new_alpha} and  Equation \ref{eq:new_value}, we will get the final equation of our model Equation \ref{eq:Fatt}. Figure \ref{fig:U-Attention Block} shows the schematic of our U-Attention Block.

\subsection{Ablation studies on the contribution of the different components of the model}\label{subsec:model-ablation}

In Subsection \ref{subsec: U-Attention origin}, we described the two important adjustments to our architecture which differentiates it from TransformerConv \cite{shi2020masked}. First, we concatenate the relative information of neighbouring node with the information of self node. Next, we replaced the normal transformer block with the U-Attention block. In this subsection, we investigate the contribution of each of these adjustments on the overall performance. Therefore, we conduct an additional ablation study where our model is compared against one that removes the U-Attention structure and another that does not concatenate information from the self-node for determining attention weights. Eliminating both adjustments collapses our model to being TransformerConv. 
The training was done on the same dataset as described in the main paper. Table \ref{table:statistic of evaluation of 4 variants of our model} shows the performance of these model variants on different vehicle/obstacle combinations.
\begin{table*}[!h]
\centering
\resizebox{\textwidth}{!}{
 \begin{tabular}{||c c c c c c c c c c||} 
 \hline
  &  & \multicolumn{4}{|c|}{success to goal rate} & \multicolumn{4}{|c|}{collision rate} \\ 
 \hline
 Num. Veh.& Num. Obs.& Our Model &  Remove Concatenation  & Remove U-Net & TransformerConv & Our Model &  Remove Concatenation  & Remove U-Net & TransformerConv \\ [0.5ex] 
 \hline\hline
1 & 0 & 1.0000 & 0.9847  & 0.9990  & 0.9894 & - & - & - & - \\
\hline
1 & 1 & 0.9677 & 0.9161  & 0.7307 & 0.7580 & 4.9616E-05 & 8.9603E-04  & 5.1616E-03  & 5.3226E-03 \\
\hline
1 & 2 & 0.8991 & 0.8109  & 0.5583 & 0.5721 & 3.3231E-04 & 2.2233E-03  & 1.1090E-02  & 1.1451E-02 \\
\hline
1 & 3 & 0.8127 & 0.7093  & 0.4677 & 0.4929 & 6.6306E-04 & 3.9955E-03  & 1.5917E-02  & 1.6024E-02 \\
\hline
1 & 4 & 0.7361 & 0.6302  & 0.3764 & 0.3922 & 1.2774E-03 & 5.4084E-03  & 2.1584E-02  & 2.1938E-02 \\
\hline
2 & 0 & 0.9984 & 0.9702  & 0.6472 & 0.6600 & 1.1006E-05 & 6.0761E-04  & 8.0570E-03  & 7.6941E-03 \\
\hline
2 & 1 & 0.9634 & 0.8688  & 0.5940 & 0.6219 & 1.6017E-04 & 3.0018E-03  & 1.0235E-02  & 9.6096E-03 \\
\hline
2 & 2 & 0.8676 & 0.7810  & 0.5425 & 0.5624 & 4.5357E-04 & 4.6598E-03  & 1.3381E-02  & 1.2796E-02 \\
\hline
2 & 3 & 0.7792 & 0.6955  & 0.5142 & 0.5209 & 6.7795E-04 & 6.4930E-03  & 1.6038E-02  & 1.6039E-02 \\
\hline
2 & 4 & 0.6781 & 0.6135  & 0.4899 & 0.4982 & 1.1135E-03 & 8.2745E-03  & 1.9067E-02  & 1.8711E-02 \\
\hline
3 & 0 & 0.9943 & 0.8107  & 0.4129 & 0.4339 & 8.8107E-05 & 3.7200E-03  & 1.4707E-02  & 1.4047E-02 \\
\hline
3* & 1* & 0.9706 & 0.7489 & 0.3950 & 0.4193 & 3.0382E-04 & 5.2493E-03  & 1.6100E-02  & 1.5655E-02 \\
\hline
3* & 2* & 0.9302 & 0.7006  & 0.3897 & 0.4174  & 5.9881E-04 & 6.6222E-03  & 1.7780E-02  & 1.7393E-02 \\
\hline
3* & 3* & 0.8903 & 0.6503  & 0.3754  & 0.4023 & 8.9677E-04 & 7.9807E-03  & 2.0166E-02  & 1.9646E-02 \\
\hline
3* & 4* & 0.8328 & 0.6161  & 0.3554  & 0.3834 & 1.6090E-03 & 9.2047E-03  & 2.2478E-02  & 2.2045E-02 \\
\hline
4* & 0* & 0.9807 & 0.6607  & 0.2894  & 0.2895 & 2.7650E-04 & 6.5573E-03  & 2.0186E-02  & 1.9967E-02 \\
\hline
4* & 1* & 0.9550 & 0.6185  & 0.2701  & 0.3048 & 5.7179E-04 & 7.8322E-03  & 2.1356E-02  & 2.0446E-02 \\
\hline
4* & 2* & 0.9279 & 0.5804  & 0.2773 & 0.2905 & 9.4375E-04 & 8.7571E-03  & 2.1877E-02  & 2.1960E-02 \\
\hline
4* & 3* & 0.8853 & 0.5426  & 0.2773 & 0.2864 & 1.4612E-03 & 1.0007E-02  & 2.3559E-02  & 2.3747E-02 \\
\hline
5* & 0* & 0.9590 & 0.5322 & 0.1890  & 0.1973 & 5.8217E-04 & 9.1856E-03  & 2.6257E-02  & 2.5964E-02 \\
\hline
5* & 1* & 0.9285 & 0.4856  & 0.1934 & 0.2078 & 1.0483E-03 & 1.0374E-02  & 2.6303E-02  & 2.6115E-02 \\
\hline
5* & 2* & 0.9037 & 0.4759  & 0.2078 & 0.2125 & 1.4063E-03 & 1.0754E-02  & 2.6261E-02  & 2.6607E-02 \\
\hline
6* & 0* & 0.9209 & 0.4128  & 0.1300 & 0.1347 & 1.1376E-03 & 1.1734E-02  & 3.1954E-02  & 3.1612E-02 \\
\hline
6* & 1* & 0.8949 & 0.3916 & 0.1419 & 0.1556 & 1.5096E-03 & 1.2351E-02  & 3.0973E-02  & 3.0917E-02 \\
\hline
6* & 2* & 0.8717 & 0.3738  & 0.1423 & 0.1479 & 1.8932E-03 & 1.2905E-02  & 3.1218E-02  & 3.1828E-02 \\
\hline
\end{tabular}
}
\caption{An ablation study showing the performance of the four variants of our model: Full Version of Our Model, Our Model without concatenation, Our Model with the U-Net structure removed and the original version of TransformerConv \cite{shi2020masked}). The metrics used for benchmarking are \emph{success-to-goal rate} and \emph{collision rate}. The evaluation is done on a completely unseen test data comprising of scenarios with 1-6 vehicles and 0-4 obstacles as shown by the corresponding rows. Each row in the table is evaluated on 4062 scenarios. The rows labeled with an asterisk (*) are those vehicle-obstacle combinations that were not even in the training set. The training set only comprised of 1-3 and 0-4 obstacles.}
\label{table:statistic of evaluation of 4 variants of our model}
\end{table*}

From Table \ref{table:statistic of evaluation of 4 variants of our model}, we notice that, adjusting the attention block to remove the U-Net architecture  but keeping the concatenation leads to a drop in performance. This demonstrates the importance of using the U-Net architecture, without which performance deteriorates. The reason is that the traditional TransformerConv architecture only uses a simple linear transform $W_{key}^{T} \cdot W_{query}$ to align the key and query. This is apparently not powerful enough to extract sufficient information when calculating attention. By adjusting the attention block to a U-Net architecture, the network in the attention block is much more powerful. So it is capable of extracting more rich and representative features which facilitates computing a more accurate attention value. 

Similar observation holds for when using only the U-Net architecture but removing the concatenation. The performance again deteriorates demonstrating the significance of using concatenation. True performance gains are only realized when both the concatenation and U-Net architecture are used together in conjunction. 

\subsection{Run Time Comparison}\label{subsec:runtime}

An advantage of our GNN model against traditional optimization based techniques is faster inference. Moreover, as the number of vehicles/obstacles is increased, the time taken by an optimization based procedure to find the correct control commands rises accordingly. However, with our graphical based architecture which allows parallel computations, the inference time remains fairly consistent. 
To demonstrate this, we run both the optimization method and our GNN model with increasing number vehicles and obstacles. For each vehicle/obstacle combination, we choose 100 cases from test dataset. 

Note that an increase in the distance between start and destination state as well as the position of the static obstacles will influence the length of the trajectory and thereby increase the prediction run time. Therefore, rather than reporting the time to complete the entire trajectory, we report the average time to complete one step towards the destination in the trajectory.

Keeping in line with the main paper, we use the same parameters for online inference here as we did for offline optimization for the purpose of label generation.  The resource we use for computation is an Intel Core i7-10750H. Table \ref{table:statistic of prediction run time} reports the run time for both the optimization based method and our model. Our model involves multiple parallel computations and can therefore be readily be deployed on a GPU too. Runtime performance of our our model on a  GeForce RTX 2070 GPU are also reported in the table.
\begin{table}
\centering
\resizebox{\columnwidth}{!}{
 \begin{tabular}{||c c c c c||} 
 \hline
 Num. Veh.& Num. Obs.& Optimization & Our model (GPU) & Our model (CPU) \\ [0.5ex] 
 \hline\hline
1 & 0 & 0.63230 & 0.00583 & 0.00408 \\
\hline
1 & 1 & 0.73012	& 0.00781 & 0.00492 \\
\hline
1 & 2 & 0.74968 & 0.00791 & 0.00522 \\
\hline
1 & 3 & 0.75063 & 0.00800 & 0.00541 \\
\hline
1 & 4 & 0.79725 & 0.00792 & 0.00568 \\
\hline
2 & 0 & 2.74806 & 0.00783 & 0.00515 \\
\hline
2 & 1 & 3.10209 & 0.00797 & 0.00570 \\
\hline
2 & 2 & 3.06539 & 0.00797 & 0.00590 \\
\hline
2 & 3 & 2.96316 & 0.00799 & 0.00634 \\
\hline
2 & 4 & 2.96338 & 0.00799 & 0.00673 \\
\hline
3 & 0 & 5.48114 & 0.00795 & 0.00591 \\
\hline
3* & 1* & - & 0.00804 & 0.00686 \\
\hline
3* & 2* & - & 0.00801 & 0.00607 \\
\hline
3* & 3* & - & 0.00799 & 0.00609 \\
\hline
3* & 4* & - & 0.00818 & 0.00635 \\
\hline
4* & 0* & - & 0.00798 & 0.00577 \\
\hline
4* & 1* & - & 0.00854 & 0.00622 \\
\hline
4* & 2* & - & 0.00864 & 0.00657 \\
\hline
4* & 3* & - & 0.00862 & 0.00679 \\
\hline
5* & 0* & - & 0.00857 & 0.00644 \\
\hline
5* & 1* & - & 0.00830 & 0.00692 \\
\hline
5* & 2* & - & 0.00886 & 0.00741 \\
\hline
6* & 0* & - & 0.00886 & 0.00732 \\
\hline
6* & 1* & - & 0.00902 & 0.00780  \\
\hline
6* & 2* & - & 0.00883 & 0.00808 \\
\hline
\end{tabular}
}
\caption{Shows the average run time per step for the optimization based procedure and our GNN model. Note that our model takes less than 10 milliseconds to get a prediction for a scenario with 6 vehicles and 2 obstacles. Meanwhile, the optimization takes more than half second (632 milliseconds) even for the simplest case with 1 vehicle and no obstacle.}
\label{table:statistic of prediction run time}
\end{table}

From the results in Table \ref{table:statistic of prediction run time}, it is evident that our model takes less than 10 milliseconds to get a prediction even for a scenario with 6 vehicles and 2 obstacles. Meanwhile, the optimization takes more than half second (632 milliseconds) even for the simplest case with 1 vehicle and no obstacle. There are two main reasons why the optimization runs much slower than our model. Firstly, the optimization procedure needs to be completed in real-time during inference, while our GNN model has already optimized its parameters during the training process. Although it takes longer to train the GNN model but once the parameters of the GNN are fixed, it is only a matter of using these pretrained weights at inference time. But for optimization, each inference is a new problem, which needs to be iteratively solved at each step and thereby accumulating the total time consumed. Besides, optimization uses a receding horizon strategy. This means in order to execute one step, we still need to predict and calculate many steps ahead, which is not an effective use of computation. But these calculations are necessary, otherwise the optimization would not be able to look ahead in order to take preemptive action to avoid collision in advance. One can reduce the prediction horizon, but it may lead to a sub-optimal trajectory.  

Another important point worth noticing, is that when the task gets more complex from 1 vehicle and 0 obstacle to 6 vehicles and 2 obstacles, the prediction run time of our GNN model only goes up marginally. But for optimization, the prediction run time increase dramatically when the number of vehicles increases. This is expected, because more vehicles means more constraints to be optimized for. Lastly, note that our model runs faster on the CPU than on the GPU. That is because during inference, the batch size is 1. Therefore, the advantage of parallelism gained from a forward pass on one sample does not compensate for the time it takes to transfer data between the CPU and GPU memory. Nevertheless, the model is still faster than compared with the optimization procedure. Moreover, GPU's are still advantageous during training wherein large batch sizes can be processed.

\subsection{Breakdown of the training data}
Note that Subsection \ref{subsec:data_collection} mentioned that the training data for which labels are generated contain between 1-3 vehicles and 0-4 static obstacles, for a total of around 20,961 trajectories. The start and destination states of the vehicles/obstacles are generated at random. Each trajectory is  collected for 120 timesteps. Therefore, the total number of scenarios generated are 2,515,320. A breakdown of this number for the different number of vehicles (1-3) and static obstacles (0-4) is shown in Table \ref{table:statistic of ground truth data}. As we saw in Subsection \ref{subsec:runtime}, increasing the number of vehicles/obstacles beyond what is enumerated in Table \ref{table:statistic of ground truth data}, significantly slows down the optimization for determining the optimal control values during this data and label generation process. Nevertheless, we demonstrated in Table \ref{table:statistic of prediction run time} that our model is still powerful enough to make fast inference for up to 6 vehicles. This is despite being trained with data that had a maximum of 3 vehicles. 

\begin{table}[!h]
\centering
\resizebox{\columnwidth}{!}{
 \begin{tabular}{||c c c c||} 
 \hline
 Num. Veh. & Num. Obs. & Num. Traj. & Num. Case \\ [0.5ex] 
 \hline\hline
 1 & 0 & 1000 & 120000 \\ 
 \hline
 1 & 1 & 1200 & 144000 \\
 \hline
 1 & 2 & 1800 & 216000 \\
 \hline
 1 & 3 & 2699 & 323880 \\
 \hline
 1 & 4 & 3289 & 394680 \\ 
 \hline
 2 & 0 & 2000 & 240000 \\ 
 \hline
 2 & 1 & 600 & 72000 \\  
 \hline
 2 & 2 & 1199 & 143880 \\  
 \hline
 2 & 3 & 1794 & 215280 \\  
 \hline
 2 & 4 & 2380 & 285600 \\  
 \hline
 3 & 0 & 3000 & 360000 \\ 
 \hline
\end{tabular}
}
\caption{Amount of  Ground Truth Data collected for 1-3 vehicles and 0-4 obstacles}
\label{table:statistic of ground truth data}
\end{table}

When running the optimization, we set the prediction horizon to be 20. The steering angle is bounded within a range of -0.8 to 0.8 radians while the pedal acceleration is bounded between -1 to 1.
Additional data is collected to make the model more robust to deviations at inference time. We simulate the vehicles diverging to random offset locations by adding noise on the position and orientation of the vehicle. After executing the first control command optimized for the the entire horizon, we add a random Gaussian noise on the position and orientation of the vehicle. The variance of the Gaussian noise decreases over time as the vehicle approaches the target position. This is to ensure that the vehicle can eventually reach its destination with minimal deviation.

\end{document}